\documentclass[runningheads]{llncs}
\usepackage{graphicx}
\usepackage{amsmath,amssymb}
\usepackage{floatrow} 
\usepackage{color}
\usepackage[width=122mm,left=12mm,paperwidth=146mm,height=193mm,top=12mm,paperheight=217mm]{geometry}
\usepackage{algorithm, algpseudocode}
\newfloatcommand{capbtabbox}{table}[][\FBwidth] 

\begin{document}
\title{A Context-aware Capsule Network for Multi-label Classification} 
\titlerunning{A Context-aware Capsule Network for Multi-label Classification}
\author{Sameera Ramasinghe\inst{1,2} \and
C.D. Athuraliya\inst{1} \and
Salman H. Khan\inst{2}}

\authorrunning{Sameera Ramasinghe, C.D. Athuraliya and Salman H. Khan}
\institute{ConscientAI Labs, Colombo, Sri Lanka \and
Australian National University, Canberra, Australia
\email{sameera.ramasinghe@anu.edu.au}}

\maketitle

\begin{abstract}
Recently proposed Capsule Network is a brain inspired architecture that brings a new paradigm to deep learning by modelling input domain variations through vector based representations. Despite being a seminal contribution, CapsNet does not explicitly model structured relationships between the detected entities and among the capsule features for related inputs. Motivated by the working of cortical network in human visual system, we seek to resolve CapsNet limitations by proposing several intuitive modifications to the CapsNet architecture. We introduce, (1) a novel routing weight initialization technique, (2) an improved CapsNet design that exploits semantic relationships between the primary capsule activations using a densely connected Conditional Random Field and (3) a Cholesky transformation based correlation module to learn a general priority scheme. Our proposed design allows CapsNet to scale better to more complex problems, such as the multi-label classification task, where semantically related categories co-exist with various interdependencies. We present theoretical bases for our extensions and demonstrate significant improvements on ADE20K scene dataset.
\end{abstract}

\section{Introduction}
\label{sec:intro}
After nearly two decades since its inception, convolutional neural networks (CNNs) \cite{Lecun98gradient-basedlearning} have eventually become the norm for computer vision tasks. Vision tasks that widely use CNNs include object recognition \cite{NIPS2012_4824,7298594}, object detection \cite{NIPS2015_5638,RedmonDGF15} and semantic segmentation \cite{long2015fully,badrinarayanan2015segnet}. Despite their popularity and high effectiveness in most vision tasks, previous works have pointed out several limitations of CNNs in vision applications. One major limitation is the notable trade-off between preserved spatial information and the transformation invariance with pooling operations. Furthermore, CNNs marginally tackle rotational invariance.

To overcome aforementioned limitations in CNNs, recently introduced Capsule Networks (CapsNets) \cite{sabour2017dynamic} propose a novel deep architecture for feature abstraction while preserving underlying spatial information. This architecture is motivated by human brain function and suggests equivariance over invariance while demonstrating comparable performance on digit classification with MNIST dataset \cite{lecun_cortes_burges_1998}. These early results of CapsNet manifest a new direction for future deep architectures. However to our knowledge, CapsNet architecture has not been used for larger and complex datasets, specifically for multi-label classification tasks where the goal is to tag an input image with multiple object categories. This is due to the reason that original CapsNet does not incorporate contextual information necessary for complex tasks such as multi-label classification. In this work we evaluate the original CapsNet architecture on a large image dataset with over 150 object classes that appear in complex real-world scenes. We then propose a new context-aware CapsNet architecture that makes informed predictions by exploiting semantic relationships of object classes as well as underlying correlations of low-level capsules. Our model is inspired by the working of human brain where contextual and prior information is effectively modeled \cite{bar2004visual}.

To enable faster training on large datasets, we \textbf{first} propose a novel weight initialization scheme based on trainable parameters with back-propagation. This update allows initial routing weights to capture low-level feature distributions and improves the convergence rate and accuracy compared to equal routing weight initialization of the original CapsNet. \textbf{Second}, we argue that the corresponding elements of primary capsule predictions are interrelated since primary capsule predictions encapsulate the attributes of object classes. In simple terms, this means that the presence of object attributes (such as position, rotation and texture) in one capsule's output are dependent on similar attributes that are detected by neighbouring capsules. This property was not utilized in the original CapsNet architecture. To characterize this, we introduce an end-to-end trainable Conditional Random Field (CRF) to encourage network predictions to be more context specific. \textbf{Third}, the original CapsNet captures the priority between primary and decision capsules independently for each data point. We argue that there exists a general priority scheme between decision and primary capsules, which is distributed across the dataset. Therefore, we propose a correlation module to capture the overall priority of primary capsule predictions throughout the dataset that effectively encapsulates broader context.

We apply proposed architecture for multi-label classification on a large scene dataset, ADE20K \cite{zhou2017scene}, and report significant improvements over the original CapsNet architecture.

\section{Related Work}
\label{relatedwork}
Hinton \textit{et al.} \cite{Hinton:2011:TA:2029556.2029562} first proposed capsule as a new module in deep neural networks by transforming auto-encoders architecture. Capsules were suggested as an alternative to widely adapted subsampling layers of CNNs and to encapsulate more precise spatial relationships. Sabour \textit{et al.} \cite{sabour2017dynamic} recently proposed a complete neural network architecture for capsules with dynamic routing and a reconstruction loss. They demonstrated state of the art performance on MNIST dataset \cite{lecun_cortes_burges_1998}. They also outperformed existing CNN architectures on a new dataset, MultiMNIST \cite{sabour2017dynamic}, created by overlaying one digit on top of another digit from a different class. More recently, Hinton \textit{et al.} \cite{e2018matrix} proposed an updated capsule architecture with a logistic unit and a new iterative routing procedure between capsule layers based on the Expectation-Maximization (EM) algorithm \cite{Dempster77maximumlikelihood}. This new capsule architecture significantly outperformed baseline CNN models on small-NORB dataset \cite{LeCun:2004:LMG:1896300.1896315}  and reported that the new architecture is less vulnerable to white box adversarial attacks. Xi \textit{et al.} \cite{xi2017capsule} extended initial CapsNet work by utilizing it on CIFAR10 classification task. However, CapsNet has not been used before for complex structured prediction tasks and our work is a key step towards this direction.

\section{Methodology}
A decision capsule is considered to be a complete representation of an object class. That means each of its scalar element describes a certain attribute of an object class such as rotation or position. These attributes may not be semantically meaningful, but an object can be completely reconstructed using the elements of the corresponding capsule. Each corresponding element of different decision capsules represents similar attributes of different objects. For example, the $i^{th}$ scalar element of $j^{th}$ decision capsule may represent the rotation of a chair, while $i^{th}$ scalar element of $(j+1)^{th}$ decision capsule may describe the rotation of a desk.

The predictions by primary capsules for decision capsules encapsulate the attributes of an object class. Therefore, the corresponding elements of outputs from primary capsules are conditioned upon each other. For example, there may be a hidden condition such that if the primary capsule is in state $A$, a chair cannot be rotated in $\alpha$ direction when a spatially nearby desk is rotated in $\beta$ direction. To exploit this behavior we feed primary capsule predictions to an end-to-end trainable CRF module to learn the inter-dependencies among attributes.

\begin{figure*}[t!]
\includegraphics[width=0.8\textwidth, trim=0cm 0.4cm 0cm 0cm]{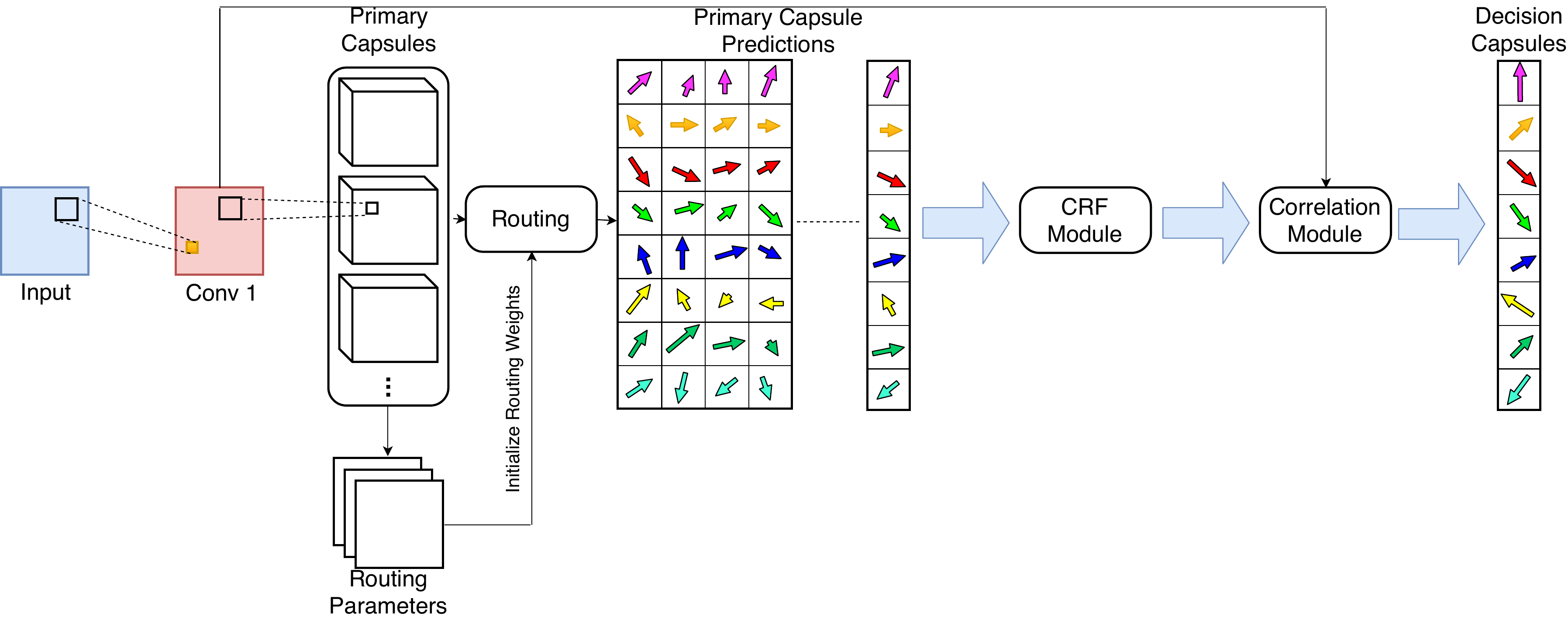}
\caption{Proposed CapsNet architecture}
\label{fig:architecture}
\end{figure*}

Here, CRF module is used as a structured prediction mechanism for each primary capsule to conditionally alter its predictions. Thus the CRF is able to capture semantic relationships across object classes. Moreover, we introduce a correlation module which can prioritize predictions by primary capsules and effectively predict decision capsules. The overall architecture is illustrated in Figure \ref{fig:architecture}. We first begin with the description of routing weight initialization and then explain the densely connected CRF and the correlation module in subsequent sections. 

\subsection{Initializing Routing Weights}
In the original CapsNet, primary capsules can be interpreted as a set of $Z$ stacked feature maps. Each primary capsule element can be considered as a part of a low-level feature. Following this assumption we rearrange primary capsules as a $N \times N \times D$ grid where $N \times N \times D$ is the total number of primary capsules. Each item in the grid is a capsule with $I$ dimensions. Hence, $D = Z/I$.

Instead of initializing routing weights equally, we modify the initial routing weights as trainable parameters and use backpropagation to train them. This forces the initial routing weights to be dependent on the low-level feature distribution resulting faster convergence.

To this end, we first define a statistical value per capsule to represent its element distribution. Let $K$ and $J$ be the number of primary and decision capsules respectively, and $C = \big\{c_1, c_2, \dots, c_K\big\}$ be the set of primary capsules. Then we map the capsules to a set $S = \big\{s_1, s_2, \dots, s_K\big\}$ where $ s_k =  \frac{\mu _{k}}{max(\sigma_{k}, \epsilon)},  \forall 0 < k < K, 0 < \epsilon << 1$ and $\mu _{k}$ and $\sigma_{k}$ are mean and standard deviation of $k^{th}$ primary capsule elements respectively. $max(\sigma_{k}, \epsilon)$ gives the maximum value between $\sigma_{k}$ and $\epsilon$ for each $k$. The operation outputs a real valued $N \times N \times D$ dimensional tensor. Treating this tensor as a stacked set of feature maps and convolving it with a single $f \times f$ kernel with $(f-1)/2$ padding, where $f$ is a positive integer (we use $f=5$ in our experiments), give a set of feature maps with dimensions $N \times N \times D$. We obtain $K \times J$ dimension matrix $B$ by transforming the elements of the feature maps as a row vector  $\hat{b} = (b_{1}, b_{2}, ..., b_{K})$, and then repeating it $J$ times. We use elements of $B$ as initial routing weights between primary and decision capsules. 
\subsection{CRF Module}
\label{crf_section}
CRF is an effective technique for structured prediction where output variables are interdependent. Furthermore, CRFs are capable of discriminative training due to conditional probabilistic modeling. They are widely used in important applications of computer vision, natural language processing and bioinformatics. We propose to use CRFs to model relationships between primary capsules in the CapsNet. The CRF models each element of each primary capsule prediction as a random variable and forms a Markov Random Field when the variables are conditioned upon inputs.

Let $P_{k,j}(i)$ denote the $i^{th}$ element of the prediction by $k^{th}$ primary capsule for the $j^{th}$ decision capsule. Considering predictions for all decision capsules by primary capsules, we define the energy function,
\begin{equation}
\label{crf}
    Z(x) = \sum_{k=0}^{K}\sum_{i=0}^{I}\sum_{j=0}^{J}E_u(P_{k,j}(i)) + \sum_{k=0}^{K}\sum_{i=0}^{I}\sum_{j'=0}^{J}\sum_{j=0,j\neq j'}^{J}E_p(P_{k,j}(i),P_{k,j'}(i))
\end{equation}
where $E_u(P_{k,j}(i))$ is cost of prediction $P_{k,j}(i)$ and $E_p(P_{k,j}(i),P_{k,j'}(i))$ is the cost of $P_{k,j}(i)$ and $P_{k,j'}(i)$ occurring simultaneously. It is evident from Equation \ref{crf} that pairwise potentials only take corresponding elements of predictions by the same primary capsule in to account. Therefore minimizing energy function in Equation \ref{crf} is equivalent to minimizing each $Z(x)_{k,i}$ for $i < I$ and $k < K$ independently where,
\begin{equation}
\label{crf1}
    Z(x)_{k,i} = \sum_{j=0}^{J}E_u(P_{k,j}(i)) + \sum_{j'=0}^{J}\sum_{j=0, j\neq j'}^{J}E_p(P_{k,j}(i),P_{k,j'}(i))
\end{equation}
Mean-field approximation provides an iterative approach to minimize dense CRF energy functions. This technique approximates a total energy function $Z(x)_{k,i}$ as a product of simple marginal energy functions $Z(X)_{k,i} = \prod_l H^l_{k,i}(x_l)$.

Zheng \textit{et al.} \cite{zheng2015conditional} leveraged this idea by formulating a dense CRF as a stack of differentiable layers. They also showed that multiple iterations of this stack of layers can be treated as an RNN. We adapt this technique to minimize the energy function \ref{crf1}. 

\kern-1.5em

\begin{algorithm}
   \caption{CRF as a stack of CNN layers}
   \label{alg:routing}
\begin{algorithmic}[1]
   \State $H_{kj}(i) = \frac{1}{X_{ik}}exp(E_u({P_{k,j}(i)})  \forall i,j,k$   \algorithmiccomment {Initialization}
   \For {$ itr=0$ \bfseries {to} $MaxItr$ }
    \State $\bar{H}_{kj}(i) = \sum_{j'} E_p  (H_{kj}(i),H_{kj'}(i))$  \algorithmiccomment  {Calculation of pair-wise potentials}
    \State $\Tilde{H}_{kj}(i) = H_{kj}(i) -  \bar{H}_{kj}(i)$  \algorithmiccomment {Addition of pair-wise potentials to unary potentials}
    \State $H_{kj}(i) = \frac{1}{X_{ik}}e^{\Tilde{H}_{kj}(i)} $  \algorithmiccomment {Normalization}
  \EndFor
\end{algorithmic}
\end{algorithm}

\kern-1.5em

The first line is the initialization. Here, $X_{i,k} = \sum_{j = 0}^{J}e^{P_{ij}^k}$ where $J$ is the number of decision capsules. Since $E_u(P_{k,j}(i))$ is the cost of the $i^{th}$ element of the prediction, we can treat the predicted element as $P_{k,j} = -E_u(P_{k,j}(i))$. This is equivalent to applying the softmax function over each set of $i^{th}$ elements of the predictions by $k^{th}$ primary capsule for $j^{th}$ decision capsules. Line number $3$ illustrates the cost of pair-wise potentials. Instead of deriving the pair-wise potential function manually, using back-propagation to find optimum mapping is both effective and efficient. Since all the corresponding element pairs have to be taken into account, we apply a fully connected layer on top of the predictions to learn  this pair-wise potential function. Since we are minimizing $Z(x)_{k,i}$ for each $i$ and $k$ independently, these layers are not connected across $i$ or $k$, which reduces the computational complexity significantly. Line number $4$ illustrates adding the unary potentials to pair-wise potentials. Line number $5$ is equivalent to applying softmax function over the outputs.

\subsection{Correlation Module}
In the CapsNet architecture, each primary capsule has a unique prediction for each decision capsule. Since primary capsules are essentially a set of low-level features, this can be viewed as each low level feature estimating the state of the output class. Moreover, each low-level feature priority depends on the output class. For example, a circle detector may perform better in predicting the state of a wheel, while a horizontal edge detector may perform better in predicting the state of a bridge.

The original routing technique tries to capture these varying priorities of primary capsules with respect to decision capsules by a weighted sum of predictions. The routing weights are adjusted in the next iteration according to the similarity between primary capsule predictions and the decision capsule of the current iteration. The magnitude of similarity is estimated by dot product. Following this method the network learns the priorities independently for each data point. However, we argue that there is also a general priority scheme that is distributed across the whole dataset, that can be learned during the training. Therefore, we propose a novel correlation based approach to discover these priorities and estimate final prediction. 

 Unlike the CRF module, our objective here is to find correlation between the attribute distributions of corresponding predictions of primary capsules and a decision capsule, instead of finding the dependency between each single corresponding attribute of predictions. Given a set of predictions for a specific decision capsule, the goal of the correlation module is to find the decision capsule elements by exploiting priorities of each primary capsule. The correlation coefficients between a decision capsule and a primary capsule predictions are learned throughout the training.  Furthermore, these correlation coefficients should depend on the low-level feature distribution and also should be trainable. To this end, we use a property of Cholesky transformation \cite{8060555} and derive a generic function to achieve this task.
 
Let two distributions be $Q$ and $R$. Cholesky transformation ensures,
\begin{equation}
\label{cholesky1}
  \left[ {\begin{array}{c}
   \bar{Q} \\
   \bar{R}\\
  \end{array} } \right]
  =
  \left[ {\begin{array}{cc}
   0 & 1\\
   \rho_1 & \sqrt{1 - {\rho_1}^2}\\
  \end{array} } \right]
    \left[ {\begin{array}{c}
   Q \\
   R\\
  \end{array} } \right]
 \end{equation}
\begin{equation}
    \bar{Q} = R, \bar{R} = \rho_1 Q + \sqrt{1 - {\rho_1}^2} R
\end{equation}
and produces two distributions $\bar{Q}$, $\bar{R}$ which are correlated by a factor of $\rho_1$. Likewise,
     \begin{equation}
\label{cholesky2}
  \left[ {\begin{array}{c}
   \bar{\bar{Q}}\\
  \bar{\bar{R}}\\
  \end{array} } \right]
  =
  \left[ {\begin{array}{cc }
   0 & 1\\
   \rho_2 & \sqrt{1 - {\rho_2}^2}\\
  \end{array} } \right]
    \left[ {\begin{array}{c}
   R \\
   Q\\
  \end{array} } \right]
 \end{equation}
\begin{equation}
     \bar{\bar{Q}} = Q, \bar{\bar{R}} = \rho_2 R + \sqrt{1 - {\rho_2}^2} Q
\end{equation}
 produces two distributions $\bar{\bar{Q}}$, $\bar{\bar{R}}$ which are correlated by a factor of $\rho_2$. Therefore if we choose,
 \begin{equation}
 \label{condition}
    \rho_2=  \sqrt{1 - {\rho_1}^2}
\end{equation}
we get $T = \bar{R} = \bar{\bar{R}}$, where $T$ and $R$ are correlated by $\rho_1$ and, $T$ and $Q$ are correlated by $\rho_2$. Using this property and considering two component distributions $D_1 = P_{k,j}$ and $D_2 = P_{k',j}$, where $P_{k,j}$ is the component distribution of $k^{th}$ primary capsule prediction for $j^{th}$ decision capsule, we obtain a new distribution $\hat{D}$, satisfying $\rho_{\hat{D}, D_1} = \rho_1$ , and $\rho_{\hat{D}, D_2} = \alpha \rho_1$. Here, $\rho_{x_1,x_2}$ denotes the correlation between the two particular distributions $x_1$ and $x_2$. Using Equation \ref{condition},
\begin{equation}
    \frac{\rho_1}{\alpha} = \sqrt{1-{\rho_1}^2}, \rho_1 = \frac{\alpha}{\sqrt{1+\alpha^2}}
\end{equation}
\begin{equation}
\label{relation}
    \hat{D} = \big[ \frac{\alpha}{\sqrt{1+\alpha^2}}D_1 + \frac{1}{\sqrt{1+\alpha^2}}D_2\big]
\end{equation}
 Using Equation \ref{relation}, we define a recursive function $f_\rho$ to obtain a correlated element distribution. 
 \begin{multline}
 \label{cholesky_big}
     f_\rho(P_{1,j}| P_{2,j}\dots, P_{k,j},\dots, P_{K,j}) = 
     \frac{\alpha_{K}}{\sqrt{1+\alpha_{K}^2}}f_\rho(P_{1,j}| P_{2,j}\dots, P_{k,j},\dots, P_{K-1,j})\\
     + \frac{P_{K,j}}{\sqrt{1+\alpha_{K}^2}}, \forall 0 < k \leq K, 0 < j \leq  J
 \end{multline}
 where $f_\rho(P_{1,j}| P_{2,j}) = \big[ \frac{\alpha_{2}}{\sqrt{1+\alpha_{2}^2}}P_{1,j} + \frac{1}{\sqrt{1+\alpha_{2}^2}}P_{2,j}\big]$.
Using this derivation, we obtain the $j^{th}$  decision capsule $C_j = f_\rho(P_{1,j}| P_{2,j}\dots, P_{k,j},\dots, P_{K,j})$.
Here, $\alpha$ requires be trainable and dependent on low-level feature distributions. Since the above operation is differentiable, the first criteria is fulfilled. To enforce $\alpha$ to be dependent on low-level features, we use the following method.
 
 Consider a $N \times N$ low-level feature map. Since we need $J(K-1)$ trainable parameters as per Equation \ref{cholesky_big}, we convolve this particular feature map with a set of  $J(K-1)$ kernels with sizes $N \times N$ each. This outputs $J(K-1)$ number of scalar values, which can be used as $\alpha$ parameters.

\section{Experiments}
We conduct experiments to demonstrate the effectiveness of each of the improvements; new initialization scheme of routing weights, CRF module and the correlation module. We use mean average precision (mAP) as the evaluation metric throughout the experiments with precision threshold 0.5. We use ADE20K dataset to evaluate the proposed architecture given its complex scenes and rich multi-label annotations for training images. ADE20K provides over $20,000$ training and testing images annotated with 150 semantic object categories.

\subsection{Importance of Trainable Initial Routing Weights}
\begin{figure}
\begin{floatrow}\CenterFloatBoxes
\ffigbox{
\includegraphics[width=0.51\textwidth, trim=0cm 0.6cm 0cm 0cm]{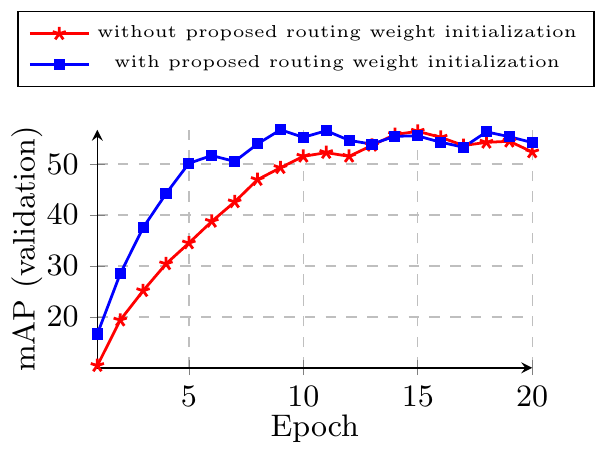}
}{
  \caption{Evaluation of convergence gain by trainable initial routing weights.}
  \label{convergence}
}
\capbtabbox{%
  \begin{tabular}{l c} \hline
  Method & mAP \\
  \hline\hline
  Original CapsNet & 42.38 \\
  Ours (RW + CRF) & 52.50 \\
  Ours (RW + CRF + CORR)  & \textbf{56.71} \\ \hline
  \end{tabular}
}
{
  \caption{Comparison of the proposed architecture with the baseline}
  \label{comparisont}
}
\end{floatrow}
\end{figure}

The goal of replacing the equal initialization of routing weights with trainable weights is faster convergence. In order to test the significance of this, we train the proposed architecture with and without the trainable initial routing scheme and test the validation mAP. The results are illustrated in Figure \ref{convergence}.

As shown in Figure \ref{convergence} the validation mAP stabilizes around $15^{th}$ epoch for the CapsNet without the proposed routing weight initialization method. On the contrary, the CapsNet with the proposed routing weight initialization method stabilizes around $9^{th}$ epoch. Therefore it is evident that the proposed method is able to achieve faster convergence compared to equal initial routing weights.

\kern-1em

\subsection{Comparison with the Baseline}
We compare the original CapsNet architecture with the proposed one by measuring the mAP measure. Table \ref{comparisont} shows the comparison results. We gain a significant $14.33$ mAP gain over total 150 object classes compared to the original architecture. Furthermore, we demonstrate performance gains by CRF and correlation modules, and show that each module provides complementary improvements. We gain an improvement of $10.12$ mAP by adding the CRF module and a $4.21$ mAP improvement by adding the correlation module on top of the CRF module. All the architectures are trained for 20 epochs.

\section{Conclusions}
In this work we attempt to overcome several limitations of CapsNet by introducing an improved architecture inspired by the contextual modeling in visual cortex \cite{bar2004visual}. Our objective is two fold: effectively capture complex interactions between primary capsules and leverage data wide correlations between representations of similar inputs. To this end, we introduced three novel ideas. Firstly, we proposed a new routing weight initialization that can be trained using back-propagation. This replaced existing equal initial routing weights with a more intuitive and efficient technique. Secondly, we introduced a CRF based method to exploit conditional attributes of primary capsule predictions to capture the context of neighbouring objects. Thirdly, we proposed a correlation module to learn dataset-wise priority scheme instead of capturing the priority separately for each data point. As demonstrated through our experiments, these improvements in CapsNet model design contributes to a substantial accuracy improvement of over 33\% in multi-label classification on a challenging dataset.

\clearpage

\bibliographystyle{splncs}

\end{document}